\documentclass[10pt,twocolumn,letterpaper]{article}

\usepackage[pagenumbers]{cvpr} 

\usepackage{graphicx}
\usepackage{amsmath}
\usepackage{amssymb}
\usepackage{booktabs}
\usepackage{paralist}
\usepackage{float}
\usepackage{paralist}

\usepackage{multirow}
\usepackage{tabularx}
\usepackage{etoolbox,siunitx}
\robustify\bfseries
\usepackage{enumitem}
\usepackage{float}
\setlist[itemize]{noitemsep, topsep=0pt}

\usepackage{comment}

\usepackage[table]{xcolor}

\usepackage{adjustbox}

\usepackage[super]{nth}
\newif\ifdrafting
\draftingtrue 
\ifdrafting
    \newcommand{\mh}[1]{{\leavevmode\color[rgb]{.2,1.0,.2}[Ming-Hsuan: #1]}}
    \newcommand{\cindy}[1]{{\leavevmode\color[rgb]{0,0.8,0.8}[Cindy: #1]}}
    
\newcommand{\ycsu}[1]{{\color{magenta}(ycsu: {#1})}}
\newcommand{\cancut}[1]{{\color{red}{#1}}}
    
\else
    \newcommand{\mh}[1]{}
    \newcommand{\cindy}[1]{}

\newcommand{\ycsu}[1]{}
\newcommand{\cancut}[1]{}

\fi

\newcommand{\ignore}[1]{}
\makeatletter
\newcommand{\printfnsymbol}[1]{%
        \textsuperscript{\@fnsymbol{#1}}%
}
\makeatother

\newcommand{\topic}[1]{\smallskip\noindent\textbf{#1.}}

\graphicspath{{figure}, {images}, {example}}

\usepackage[pagebackref,breaklinks,colorlinks,bookmarks=false,citecolor=blue,linkcolor=blue]{hyperref}

\usepackage[capitalize]{cleveref}

\crefname{section}{Sec.}{Secs.}
\Crefname{section}{Section}{Sections}
\Crefname{table}{Table}{Tables}
\crefname{table}{Tab.}{Tabs.}

\def\cvprPaperID{xxxx} 
\def\confName{ICCV}
\def\confYear{2023}

\begin{document}

\title{
Video Generation Beyond a Single Clip
}

\author{Hsin-Ping Huang$^{1,2}$, Yu-Chuan Su$^1$, Ming-Hsuan Yang$^{1,2,3}$\vspace{1mm}\\
$^1$Google Research~~~$^2$UC Merced~~~$^3$Yonsei University\\\vspace{1mm}}

\twocolumn[{
\renewcommand\twocolumn[1][]{#1}
\maketitle
}]

\begin{abstract}
    
We tackle the long video generation problem, i.e.~generating videos beyond the output length of video generation models. Due to the computation resource constraints, video generation models can only generate video clips that are relatively short compared with the length of real videos. Existing works apply a sliding window approach to generate long videos at inference time, which is often limited to generating recurrent events or homogeneous content. To generate long videos covering diverse content and multiple events, we propose to use additional guidance to control the video generation process. We further present a two-stage approach to the problem, which allows us to utilize existing video generation models to generate high-quality videos within a small time window while modeling the video holistically based on the input guidance. The proposed approach is complementary to existing efforts on video generation, which focus on generating realistic video within a fixed time window. Extensive experiments on challenging real-world videos validate the benefit of the proposed method, which improves over state-of-the-art by up to 9.5\% in objective metrics and is preferred by users more than 80\% of time.

\end{abstract}

\vspace*{-0.05in}
\section{Introduction}
\vspace*{-0.05in}
\label{sec:intro}

\begin{figure}[ht]
    \centering
    \includegraphics[width=0.5\textwidth]{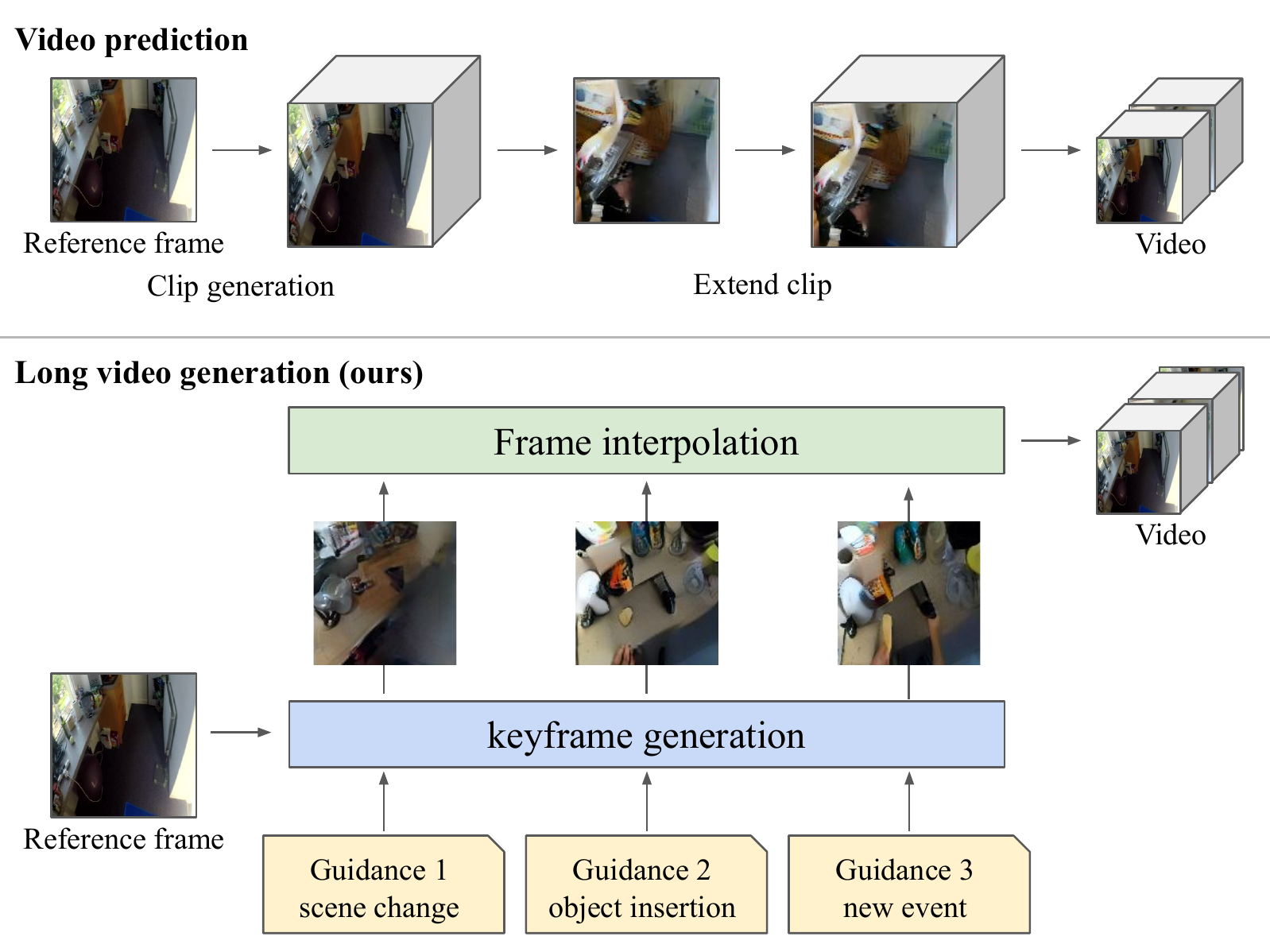}
    \vspace{-15pt}
    \caption{
        \textbf{Long video generation.} Video generation models can only generate a relatively short video clip. Existing works~\cite{ge2022long} generate long videos using a sliding window approach that extends the previously generated video clip, which often leads to videos with repetitive patterns. Instead, we study the problem of generating long videos using additional guidance, which helps to generate videos covering multiple, non-recurrent events. We also propose a two-stage approach for the long video generation problem, which is complementary to existing efforts on video generation.
    }
    \vspace{-12pt}
    \label{fig:teaser}
\end{figure}

Video generation has recently attracted increasing attention. As a natural extension of image generation, most existing works treat video generation as a 3D volume prediction problem and focus on generating realistic video clips. While this paradigm has demonstrated impressive progress in a wide range of video generation tasks such as frame prediction~\cite{srivastava15_unsup_video,finn2016Unsupervised,amersfoort2017transformationbased,Oh2015action,mathieu2015deep}, class-conditional generation, and unconditional video generation~\cite{carl2016,tgan2017,tulyakov2018MoCoGAN,TGAN2020,Clark2019AdversarialVG}, the length of the video clip that can be generated is inevitably limited by the computation resource. Due to the significant computation and memory overhead incurred by state-of-the-art video generation models, they often generate video clips that are much shorter than real videos.

In order to generate long videos that match the length of real videos, existing works adopt a sliding window approach. Specifically, the model generates one video clip at a time in temporal order while taking the previously generated frames as input. The previously generated frames serve as the condition for the model to ensure that the generated video is consistent across video clips. This approach has been successfully applied to generate long videos~\cite{ge2022long}. However, the results are far from satisfactory, particularly in real data domains. The synthesized videos are often limited to homogeneous videos of natural scenes and videos with recurrent human actions. In contrast, many real videos contain dynamic scenes with multiple, non-recurrent events.

The existing sliding window approach is sub-optimal from two aspects. First, it implicitly assumes that video generation is a Markov process, where a new clip only depends on the previous clip or even the last frame of the previous clip. This is not true in many videos, e.g.~an object may leave and re-enter the video which introduces a long-range dependency to the video. Second, it assumes that the initial frames are a sufficient condition for generating the video clip. However, it is well known that there may be multiple valid futures given the initial condition for a video~\cite{Yushchenko2019MarkovDP}, and it is unlikely to infer the correct future purely from the initial frames. 
To overcome these problems, it is essential to use additional guidance to control the generation process of long videos.  

In this work, we propose to tackle the \emph{long video generation} problem. Given a video clip generation model, a series of guidance, and a reference frame representing the initial condition of the video, the goal is to generate a long video that is beyond the capability of the clip generation model. In this work, we choose to use object labels as the input guidance, where the guidance describes the set of objects that will appear in the video clip. While other types of guidance are also possible, we choose object labels because they can be provided by users easily and naturally supports video manipulation such as content insertion or removal. The long video generation problem aims to extend the ability of existing video generation models to generate realistic video covering diverse content and multiple events and is complimentary to existing efforts that focus on generating high-quality videos within a fixed temporal window.

To solve the long video generation problem, we propose a two-stage approach by decomposing the problem into a keyframes generation problem followed by a frame interpolation problem. We first predict all the keyframes jointly based on the input guidance and reference frame. These keyframes represent the starting frames of each video clip. We then generate the entire video by predicting the intermediate frames between keyframes, using the video clip generation model. See Fig.~\ref{fig:teaser}. The two-stage approach allows us to utilize existing video generation models that are highly optimized and can generate high-quality, realistic videos within a short temporal window. It also allows us to model the full video jointly and capture long-range dependencies through keyframe generation. Our evaluations show that the holistic keyframe modeling help to maintain consistency throughout the video.

We conduct extensive quantitative and qualitative studies on both real and synthetic data. In particular, we evaluate our method on the EPIC Kitchen dataset, which is challenging for video generation due to the rapid motion and complex scenes. Empirical results verify the advantage of the proposed long video generation method and show that it outperforms state-of-the-art video generation models by 9.5\% on LPIPS.

Our main contributions are as follows. First, we study the long video generation problem which aims to extend the capability of existing video generation models to generate long videos with dynamic senses and non-recurrent events. Second, we propose a two-stage approach for the long video generation problem. Finally, we conduct extensive evaluations to validate the efficacy of our proposed framework.

\vspace*{-0.05in}
\section{Related Work}
\vspace*{-0.05in}
\label{sec:related}

\topic{Video synthesis}
Video prediction, class-conditional video generation and unconditional video generation have been widely studied as the sub-tasks of video generation. 
GAN-based methods~\cite{carl2016,tgan2017,tulyakov2018MoCoGAN,Clark2019AdversarialVG,TGAN2020,stylegan-v,tian2021a,brooks2022generating} have demonstrated early success to generate short video clips, while the generation quality decreases significantly when applying to long videos.
Diffusion models~\cite{voleti2022MCVD,yang2022diffusion,ho2022video} are recently introduced for video generation. However, the slow sampling speed of diffusion models limits their ability to generate long videos.
Auto-regressive models are first developed to synthesize raw pixels in videos~\cite{nal2017video,Weissenborn2020Scaling,babaeizadeh2021fitvid}. Thanks to the development of vector quantization~\cite{esser2020taming,oord2017vqvae} and transformer~\cite{devlin2019bert} models, auto-regressive methods are adapted to predict discrete tokens in the latent space~\cite{yan2021videogpt,lemoing2021ccvs,ge2022long} with impressive visual quality.
In this work, we build upon recent advances of VQVAE and non-autoregressive transformer models~\cite{chang2022maskgit,yu2022magvit}.

\begin{figure*}[t]
    \centering
    \includegraphics[width=.85\textwidth]{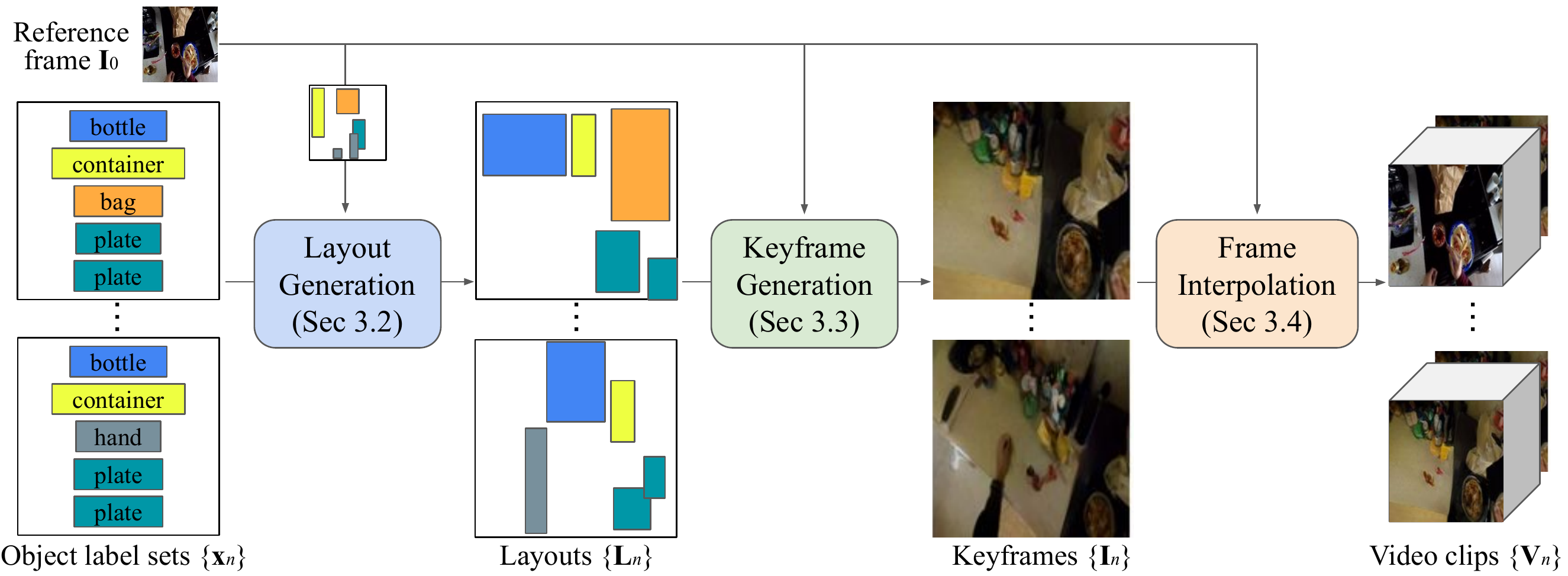}
    \vspace*{-3pt}
    \caption{\textbf{Approach overview.} The proposed method consists of three stages: 1) generating the series of layouts from the series of object label sets and a reference first frame image, 2) synthesizing the keyframes of the video from the predicted layout sequence and the reference image, and 3) interpolating the synthesized keyframe sequence to obtain the complete video.
    }
    \vspace{-9pt}
    \label{fig:overview}
\end{figure*}

Despite the recent success in video generation, these models mostly focus on generating short clips (\eg 16-frame videos) and are limited to synthesizing videos in specific domains~\cite{Villegas2022PhenakiVL,ge2022long}, such as human actions~\cite{soomro2012ucf,Siarohin_2019_NeurIPS}, sky timelapse~\cite{Xiong_2018_CVPR}, robotics videos~\cite{ebert2017bair}.
Most recently, a few text-to-video models~\cite{wu2021godiva,nuwa2022wu,hong2022cogvideo,singer2023makeavideo,ho2022imagen} are developed to generate videos given natural language inputs. However, these models are limited to videos with single scenes without a meaningful storyline.
In contrast, our work aims to generate videos with diverse content and novel events.

There are few works exploring complex video generation conditioned on input guidance at multiple timesteps while limited to synthetic environments~\cite {bar2020compositional,yu2022modular,ge2022long}. Our work focuses on the real-world dataset.

\topic{Story visualization and image manipulation}
Story visualization~\cite{li2018storygan,song2020CPCSV,adyasha2021improving,maharana2021integrating} focuses on synthesizing a sequence of images that visualize a story of multiple sentences. Each sentence in the story corresponds to one synthesized image. 
GeNeVA~\cite{el2019geneva,fu2020sscr,Liu_2020_ACMMM,cong2022lsgan} is a conditional text-to-image generation task developed on CoDraw~\cite{kim2019codraw} dataset. It studies the problem of constructing a scene iteratively based on a sequence of descriptions.
However, these two lines of work are limited to experiments on synthetic and cartoon data. These approaches focus on generating a few frames of a visual story instead of videos.
In addition, the inputs to these methods are natural language descriptions which might exist in ambiguity and do not clearly describe the objects in the image, unlike object labels.
In contrast, we focus on experiments on real-world data and generating videos given a series of object labels as inputs.

\vspace*{-0.05in}
\section{Approach}
\vspace*{-0.05in}
\label{sec:approach}

In this section, we introduce the proposed method for long video generation.
We first give an overview of the approach and then describe the three main components.

\subsection{Overview}
We define our problem as follows. The model takes 1) a series of $N$ sets of object labels $\{\mathbf{x}_1, \mathbf{x}_2, \cdots, \mathbf{x}_N\}$, and 2) a reference frame, $I_{0}$ as input.
The reference frame is the starting frame of the video and is the same initial condition as existing video prediction task.
The object label sets serve as the guidance for video generation.
Each object label set $\mathbf{x}_n$ contains $k^{n}$ object labels which indicates the objects that will appear in the video.
Note that $k^{n}$ may vary across different timestep $n$.
Our goal is to synthesize a video $\mathbf{V}$ that covers the content provided in the guidance while maintaining the same style as the initial frame $\mathbf{I}_0$.

We present the overview of the long video generation framework in \cref{fig:overview}. 
To tackle the long video generation problem, our core idea is to model the entire video jointly through keyframe generation and utilize existing models to generate high-quality videos within a short temporal window.
We achieve these by 1) generating a series of $N$ keyframes in the video given the object label sets, each keyframe presents the starting frames of each video clip, and 2) predicting the intermediate frames between adjacent keyframes to obtain the complete video. 
To reduce the difficulty of keyframe generation, we use layouts as the intermediate representation to bridge the gap of generating 2D images from symbolic object labels~\cite{Hong_2018_CVPR,johnson2018image}.
We introduce an additional stage of layout generation that creates an explicit 2D structure of the scene as an intermediate input to constrain the keyframe generation. 
It brings an additional benefit that the users can manipulate the generated videos by editing the layouts such as content insertion or removal.
The proposed approach consists of three steps: layout generation, keyframe generation, and frame interpolation.

\subsection{Layout Generation}
\label{sec:3_1}

We first generate a series of layouts from the object label sets to explicitly constrain the keyframe generation.
Given a series of $N$ object label sets $\{\mathbf{x}_n\}$, and a reference first frame $\mathbf{I}_0$, we synthesize a series of layouts $\{\mathbf{L}_1, \mathbf{L}_2, \cdots, \mathbf{L}_N\}$, which represents the layouts of the $N$ keyframes in the video. 
Since the reference first frame $\mathbf{I}_0$ is given, we assume $\mathbf{L}_0$ is known as well and can be used as a reference layout to synthesize $\{\mathbf{L}_1, \mathbf{L}_2, \cdots, \mathbf{L}_N\}$.
We define the layout $\mathbf{L}$ as a set of bounding boxes with variable length $k^n$, \ie $\{\mathbf{b}_1, \mathbf{b}_2, \cdots, \mathbf{b}_{k^n}\}$. 
The bounding box contains the attribute of its object label, x-coordinate of the center, y-coordinate of the center, width and height, \ie $\mathbf{b}=\{\mathbf{c}, \mathbf{x}, \mathbf{y}, \mathbf{w}, \mathbf{h}\}$.

Our layout generator is motivated by BLT~\cite{kong2022blt}. 
We first apply tokenization to encode the object label sets $\{\mathbf{x}_n\}$ into discrete tokens. We then learn a transformer model that predicts the layout tokens given the object label sets as input. The ground truth tokens are obtained from tokenizing the layouts $\{\mathbf{L}_n\}$.
Though the object label sets and layouts have variable length $k^n$, we pad the token sequences to fixed length in practice.
Specifically, the input series of $N$ object label sets with $k^n$ labels in each set can be flattened into a sequence of tokens.
Similarly, the layout sequences can be flattened into token sequences as well, 
where the values of the bounding box attributes are simply discretized by uniform quantization~\cite{kong2022blt}.

The tokens of the reference layout $\mathbf{L}_0$ and the object labels $\mathbf{c}$ are known and the other attributes of the bounding boxes $\{\mathbf{x}, \mathbf{y}, \mathbf{w}, \mathbf{h}\}$ are replaced with a [MASK] token. The transformer model is trained to predict the missing tokens for the layout series $\hat{\mathbf{L}}_1-\hat{\mathbf{L}}_N$~\cite{kong2022blt}.
Different from BLT~\cite{kong2022blt} which takes a single object label set as input and predicts a single layout, our model takes the reference layout and a series of object label sets as input, and predicts multiple layouts at the same time. Thus, our model preserves the temporal consistency of the layout sequences while BLT fails to.

\subsection{Keyframe Generation}
\label{sec:3_2}

Next, we generate a sequence of keyframes from the predicted layout sequences. 
Given a reference first frame $\mathbf{I}_0$ and a series of layouts $\{\hat{\mathbf{L}}_1, \hat{\mathbf{L}}_2, \cdots, \hat{\mathbf{L}}_N\}$ synthesized in the previous stage, we aim to generate a sequence of keyframes $\mathbf{I}_1$-$\mathbf{I}_N$.
Each pair of adjacent keyframes will be used as the start and end frame to generate one video clip out of the $N$ clips in the entire video.
Following~\cite{chang2022maskgit}, we convert the keyframe generation into tokenization and sequence prediction. We use a VQVAE~\cite{esser2020taming} model to encode the raw image pixels $\mathbf{I}$ into discrete visual tokens $\mathbf{e}$, and a bidirectional transformer model is used to predict the masked image tokens \ie target keyframes.
Finally, the decoder of VQVAE is used to map the visual token $\hat{\mathbf{e}}$ into raw images $\hat{\mathbf{I}}$.

Specifically, the input series of layouts $\{\mathbf{L}_0, \hat{\mathbf{L}}_1, \cdots, \hat{\mathbf{L}}_N\}$ are flattened into a sequence of discrete tokens. The reference keyframe $\mathbf{I}_0$ and the target keyframe sequence $\mathbf{I}_1$-$\mathbf{I}_N$ are transformed into discrete visual tokens and flattened into sequences as well, \ie $\{\mathbf{e}_0, \mathbf{e}_1, \cdots, \mathbf{e}_N\}$. 
At training time, the tokens of input layouts and the visual tokens of the target keyframes are concatenated. The tokens are randomly masked out and the transformer model is trained to predict the missing token. 
Specifically, given a sequence $
\mathbf{s}=\{\mathbf{e}_0, \mathbf{L}_0, \hat{\mathbf{L}}_1, \cdots, \hat{\mathbf{L}}_N, \mathbf{e}_1, \cdots, \mathbf{e}_N\}$ in dataset $\mathcal{D}$, we replace the tokens in the sequence with the [MASK] token and obtains the masked sequence $\mathbf{s}_M$.
We minimize the negative log-likelihood of predicting the masked tokens $\mathbf{s}_t, t \in M$. 

\begin{equation}
    \mathcal{L} ={-}\mathop{\mathbb{E}}\limits_{ s \in \mathcal{D}}  \sum_{t\in M} \log p(s_{t}|s_M) 
\end{equation}

\begin{figure}[t]
    \centering
    \includegraphics[width=0.5\textwidth]{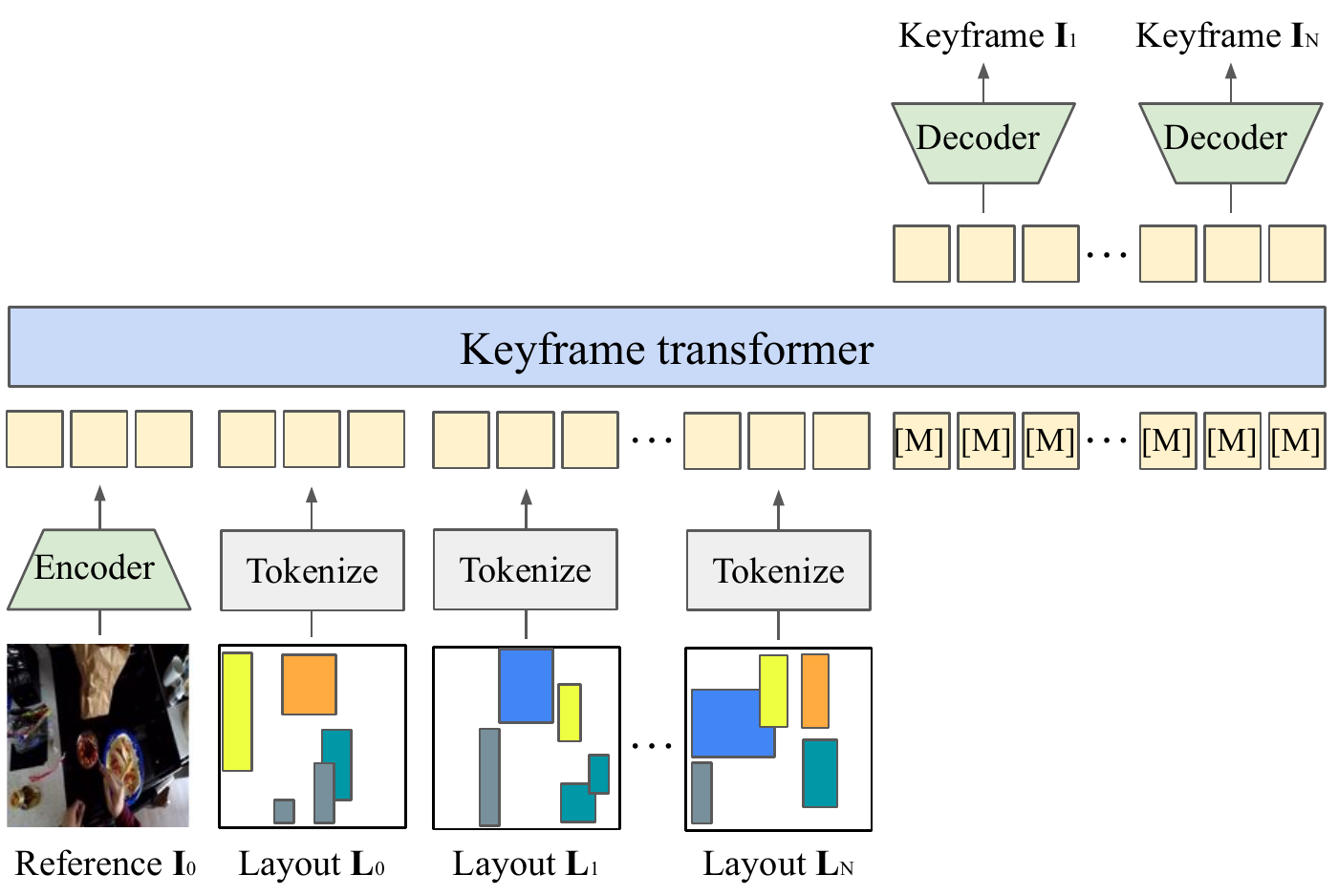}
    \vspace{-18pt}
    \caption{\textbf{Keyframe generation.} The reference frame and predicted layouts are first converted into discrete tokens. A transformer takes the tokens as input and predicts the tokens of the keyframes, which are then decoded into the final keyframes. Note that the transformer generates all keyframes jointly, which provides a more holistic model for the video.
    }
    \vspace{-9pt}
    \label{fig:transformer}
\end{figure}

At test time, the tokens of layout sequence and the first frame $\{\mathbf{e}_0, \mathbf{L}_0, \hat{\mathbf{L}}_1, \cdots, \hat{\mathbf{L}}_N\}$ are given, and the model predicts the tokens of the following keyframes $\hat{\mathbf{e}}_1-\hat{\mathbf{e}}_N$.
Finally, the decoder of VQGAN is used to reconstruct the target keyframes $\hat{\mathbf{I}}_1-\hat{\mathbf{I}}_N$.
\cref{fig:transformer} shows our keyframe generation approach.

Compare with~\cite{chang2022maskgit}, our keyframe generation method generates all the frames jointly. This provides a more holistic model for the entire video. As we show in the experiment, this helps to improve the consistency across keyframes and improves the coherency of the video.

\subsection{Frame Interpolation}
\label{sec:3_3}

Finally, given the reference first frame $\mathbf{I}_0$ and a sequence of generated keyframes $\mathbf{\hat{I}}_1 \ldots \mathbf{\hat{I}}_N$, we apply an existing video generation model to generate the complete video. 
In particular, we use MAGVIT~\cite{yu2022magvit} to generate intermediate frames following the video interpolation task of the original model.
Specifically, the model takes the initial and final frame of the video as input.
It first converts the input frames into discrete video tokens using 3D-VQVAE.
A transformer model is then used to predict the tokens of intermediate frames.
Finally, the interpolated video tokens are mapped back to the raw videos by the 3D decoder.

We re-train the video generation model on our data. During inference time, given two consecutive keyframes $\mathbf{\hat{I}}_{n-1}$ and $\mathbf{\hat{I}}_{n}$, the model predicts the video token sequences that connect between the two keyframes, \ie $\mathbf{\hat{z}}_n$.
We predict $N$ clips of video tokens $\mathbf{\hat{z}}_{1}, \mathbf{\hat{z}}_{2}, \cdots, \mathbf{\hat{z}}_{N}$. Finally, the video token sequences are concatenated and then mapped to the raw video pixels $\mathbf{V}$ by the 3D decoder.

\vspace*{-0.05in}
\section{Experiments}
\vspace*{-0.05in}
\label{sec:experiments}

We validate our approach on both real and synthetic data. We first evaluate the video generation results on challenging real-world videos. Next, we study the keyframe generation results on both real and synthetic data.

\topic{Dataset}
We validate our method on two datasets.

\textit{EPIC Kitchen}~\cite{Damen2022RESCALING} is a real video dataset consisting of egocentric videos of kitchen activities. It contains 700 videos with a total length of 100 hours. Compared with other commonly used datasets for video generation research, e.g.~UCF~\cite{soomro2012ucf}, Kinetics~\cite{kay2017kinetics}, and BAIR~\cite{ebert2017bair}, the content of EPIC Kitchen videos are more dynamic. The objects move in and out of the camera field-of-view frequently, and the scene and camera viewpoint may change rapidly. To synthesize such video, the video generation model needs to generate multiple, non-recurrent events within a short time window. This aligns with our goal for long video generation, and EPIC Kitchen serves as an ideal test bed for the problem and approach. Therefore, we choose EPIC Kitchen over other datasets for our experiments. 

We preprocess the EPIC Kitchen videos as follows. We follow the original train and test split and cut the videos into 64-frame sequences. We first re-sample the sequences with double the frame rate, so that each 64-frame sequence covers a 5-second video with the five keyframes sampled equidistantly. This leads to 276k sequences for training and 3k sequences of non-overlapping frames for test. 
We use MaskFormer~\cite{cheng2021maskformer} to extract the semantic map for each frame and convert them into the object labels and bounding boxes, which serve as the input guidance and ground truth layout. Please refer to the supplementary material for details.

\textit{CoDraw}~\cite{kim2019codraw} is a synthetic dataset consisting of virtual scenes made by clip objects. It contains 10k scenes consisting of 58 different objects. Each scene comes with a sequence of images showing the step-by-step construction process of the scene. While it is not a video dataset, it contains diverse object classes and provides complete annotation for the object labels and layouts, which is ideal for controlled experiments. We use the sequence of images as video keyframes and evaluate keyframe and layout generation on CoDraw. 
To analyze the temporal consistency between the generated keyframes, we extend the original CoDraw data by creating 6 different appearances for each clip object class and re-render the data using the original layouts.
This resulting dataset consists of 67k training sequences and 4k test sequences.

\begin{table}[t]
    \small
    \centering
    \tabcolsep=0.12cm
    \caption{\textbf{Quantitative results for video generation.} 
        We report the metrics on the EPIC Kitchen dataset~\cite{Damen2022RESCALING}. Image metrics are averaged across all video frames. Our method consistently outperforms state-of-the-art video generation methods.
    }
    \vspace{-6pt}
    \label{tab:video} 
    \begin{tabular}{lcccc}
        \toprule
        Methods & FVD~$\downarrow$ & PSNR~$\uparrow$ & SSIM~$\uparrow$ & LPIPS~$\downarrow$ \\
        \hline
        MAGVIT (frame pred.)~\cite{yu2022magvit}  & 421.7 & 11.316 & 0.175 & 0.482 \\
        MAGVIT (class cond.)~\cite{yu2022magvit} & 400.9 & 11.398 & 0.178 & 0.476 \\
        \midrule
        Ours                   & 380.8 & 12.037 & 0.206 & 0.431 \\
        Ours (GT Layouts)   & 363.5 & 13.566 & 0.287 & 0.350 \\
        Ours (GT Keyframes) & 314.7 & 15.591 & 0.386 & 0.253 \\
        \bottomrule
    \end{tabular}
    \vspace{-12pt}
\end{table}

\begin{figure}[t]
    \centering
    \includegraphics[width=\linewidth]{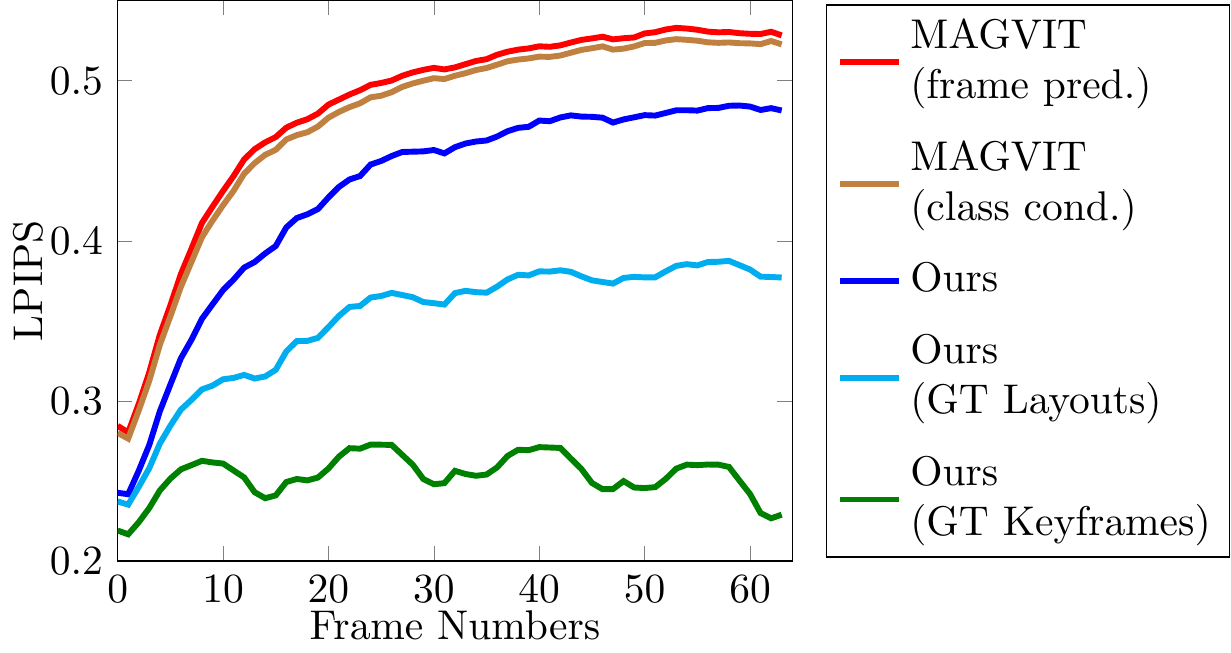}
    \vspace{-18pt}
    \caption{\textbf{Per-frame results for video generation.} 
        X-axis shows the frame number, ranging from 0 to 64. Y-axis shows the LPIPS of a specific frame averaging over all test videos.
        The performance of MAGVIT drops rapidly over time, while our approach can slow down the quality degradation. Figures are best viewed in color.
    }
    \label{fig:video_plot}
    \vspace{-9pt}
\end{figure}

\topic{Evaluation metrics} We use the following metrics for performance evaluation: 
\begin{itemize}[leftmargin=*,label=$\bullet$,topsep=2pt]
    \setlength{\itemsep}{2pt}
    \setlength{\parskip}{2pt}
    \item Fr\`{e}chet Video Distance (FVD)---assesses the quality of generated videos. Specifically, it measures whether the distribution of generated videos is close to that of real videos in the feature space. Following the original paper, we use I3D model trained on Kinetics-400 for video features. 
    \item Fr\`{e}chet Inception Distance (FID)---assesses the quality of generated images, similar to FVD. We use inception V3 for image features. FID is used to evaluate the keyframe quality.
    \item Learned Perceptual Image Patch Similarity (LPIPS)---assesses the perceptual similarity between the generated video frames and the ground truth video frames. We compute these metrics for every frame and report the average.
    \item PSNR, SSIM---assess the similarity between the generated frames and ground truth frame, similar to LPIPS.
\end{itemize}

\topic{Implementation details}
In our experiments, each video sequence contains 64 frames, and the videos are generated at 128$\times$128.
We sample one keyframe every 16 frames in the 64-frame clips so that the keyframe sequence contains $N=4$ synthesized keyframes at 256$\times$256.
More implementation details are in the supplementary materials.

\begin{figure*}[t]
    \centering
    \includegraphics[width=1.0\textwidth]{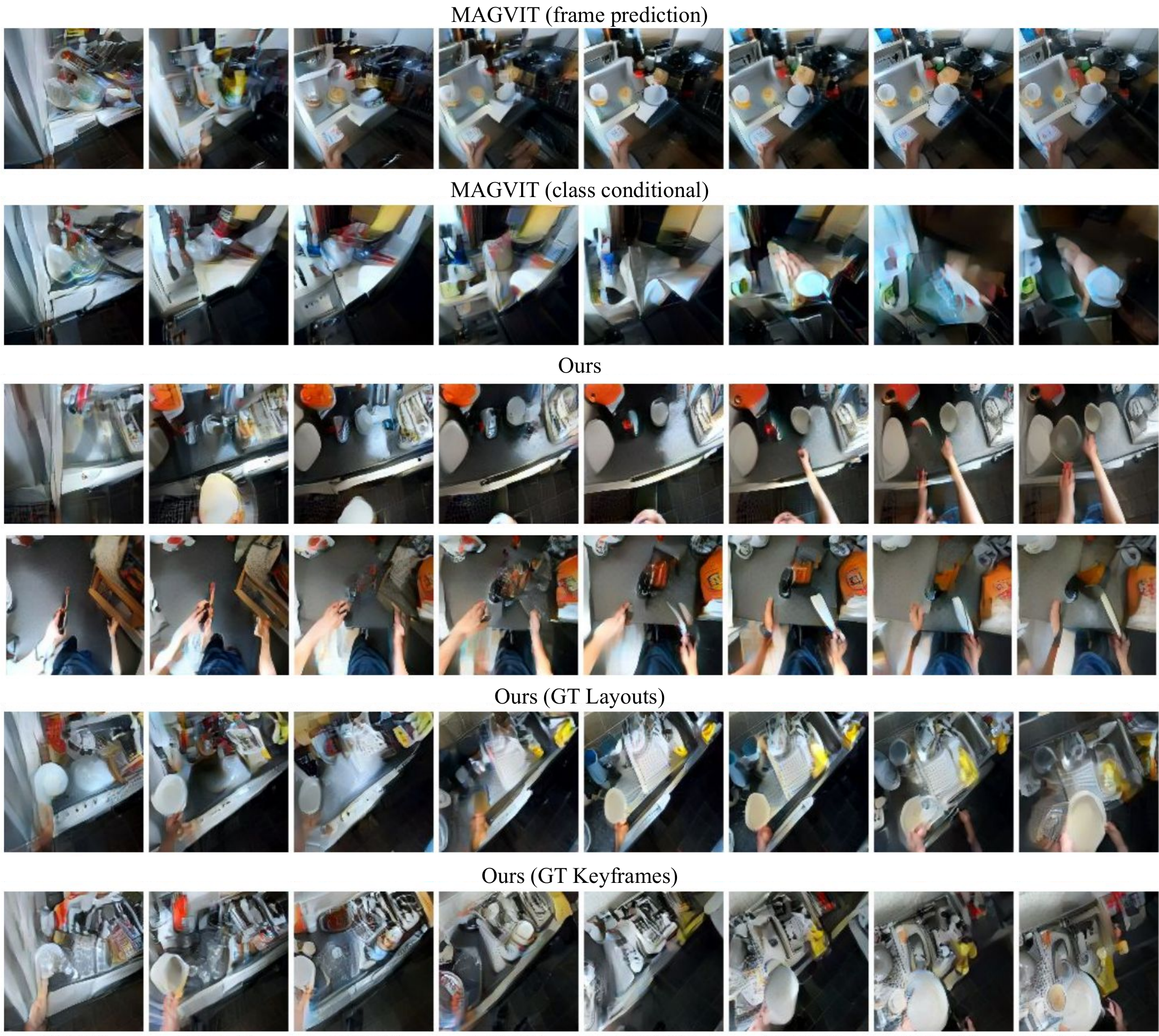}
    \vspace{-18pt}
    \caption{\textbf{Visual results for video generation.} MAGVIT (frame pred./class cond.) generates video with homogeneous contents, and the quality degrades for long sequences.
    Our method generates video with scene change (floor to table), object deletion (utensils on left hand) and object insertion (knife on right hand), showing the ability of our model to generate videos with multiple events.
    We further show that Our (GT Layouts) generates videos close to the upper bound results of Our (GT Keyframes). %
    On the other hand, Ours generates videos that match the content but with different layouts.
    The results validate the ability of our model to generate videos that satisfy different levels of the input guidance \ie object label sets or (more constrained) layouts.
    }
    \label{fig:video}
    \vspace{-12pt}
\end{figure*}

\begin{figure}[t]
    \centering
    \includegraphics[width=0.5\textwidth]{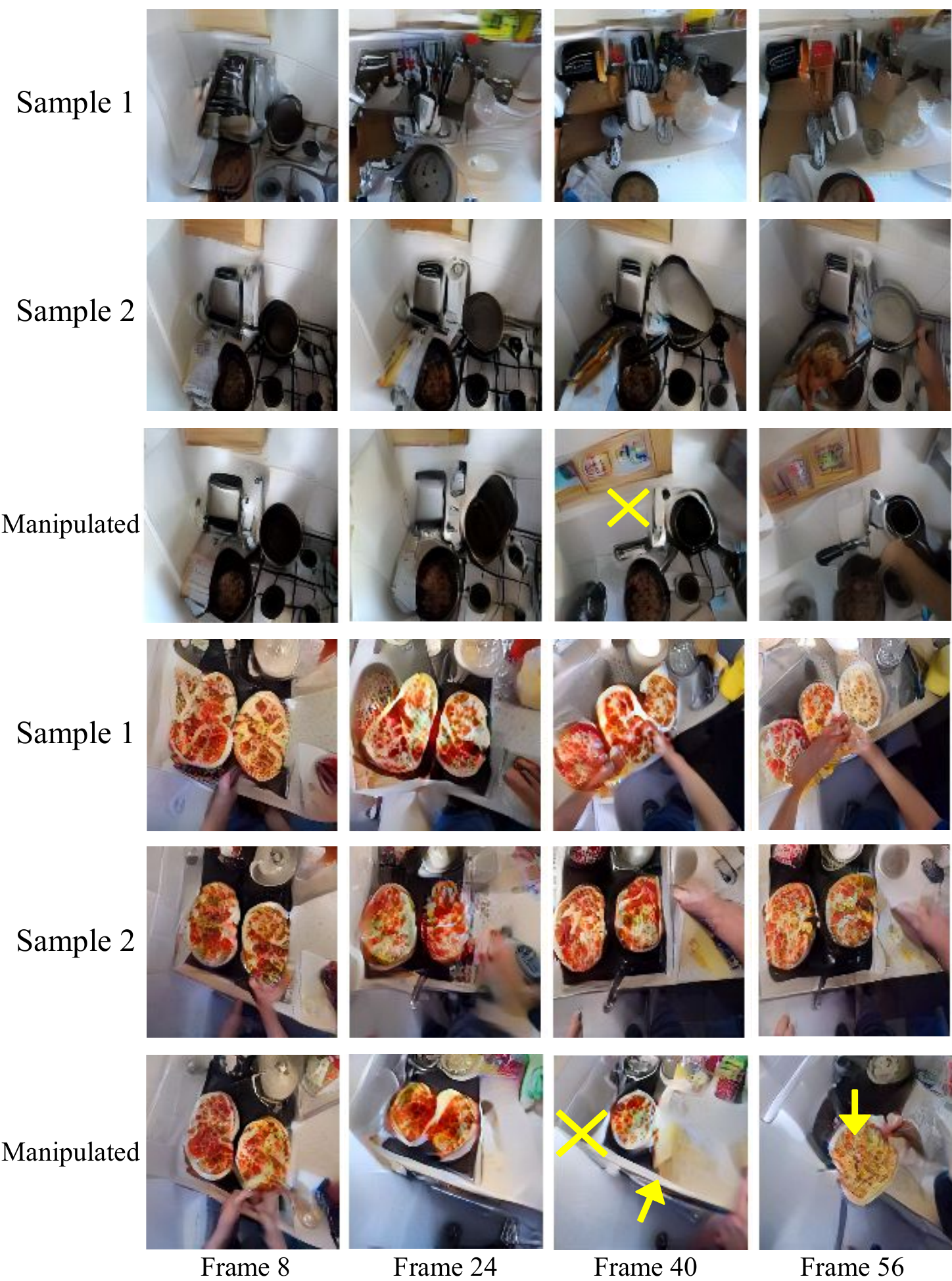}
    \vspace{-15pt}
    \caption{
    \textbf{Video manipulation through layouts.} Users can generate different videos by sampling different layouts, as shown in the first two rows. Users can even manipulate the videos by editing the layouts. The yellow cross in the third row shows object deletion, and the yellow arrow in the last row shows object movement.
    }
    \vspace{-9pt}
    \label{fig:video_our2}
\end{figure}

\subsection{Video Generation}
\label{sub:video}

First, we evaluate the video generation results on the EPIC Kitchen dataset. The goal is to verify the effectiveness of additional guidance and holistic video modeling in long video generation.

\topic{Baselines} We compare with the following state-of-the-art video generation methods
\begin{itemize}[leftmargin=*,label=$\bullet$,topsep=2pt]
    \setlength{\itemsep}{2pt}
    \setlength{\parskip}{2pt}
    \item MAGVIT (frame prediction)~\cite{yu2022magvit}: given the reference frame as input, we apply MAGVIT to generate a 16-frame clip. We then take the last predicted frame as input to iteratively generate the entire video. This is the standard sliding window approach for long video generation.
    \item MAGVIT (class conditional)~\cite{yu2022magvit}: we condition the MAGVIT model on both the reference frame and the object label. This extends the sliding window approach to take additional guidance similar to our method.
\end{itemize}
To understand how the quality of layout and keyframe generation affects the video generation results, we also compare with two variants of our method that take the ground truth layouts and keyframes as inputs respectively.

\topic{Quantitative results}
The results are in \cref{tab:video}. Our method consistently outperforms the baselines in terms of video quality, and the generated videos are closer to the ground truth videos. The results verify the effectiveness of the proposed approach. Note that MAGVIT (class conditional) performs better than MAGVIT (frame prediction), which shows the benefit of additional guidance. Our approach further improves MAGVIT (class conditional), which suggests that both the additional guidance and the holistic video modeling provided by our approach are helpful for long video generation.

On the other hand, Ours (GT Layouts) and Ours (GT Keyframes) significantly improve the output video quality. The results suggest that long video generation has significant room for improvement given the very same video generation model, and improving the video generation model alone may not be sufficient for solving the video generation problem. It verifies the importance of the long video generation problem. Note that our multi-stage approach allows users manually improve the intermediate representations, e.g.~provide more detailed layouts, which allows further improvement for the generated videos in an interactive generation setting.

\cref{fig:video_plot} shows the per-frame LPIPS score. The generated image quality degraded rapidly in MAGVIT, especially when we try to generate videos beyond the length of the training clip (i.e.~16 frames). In contrast, our approach experiences a slower quality degradation, which further verifies its benefit in generating video beyond the training clip.

\topic{Qualitative results}
\cref{fig:video} show the qualitative results. MAGVIT tends to predict a relatively static video, which is consistent with the observations in prior works. On the other hand, our method generates video with scene change (wall to table), object deletion (plate at the bottom) and object insertion (left/right hand), showing the ability of our model to generate videos with multiple events. Compared with the upper bound results of Ours (GT Keyframes), our method generates videos that match the content but with different layouts or locations.

\cref{fig:video_our2} shows that our method allows users to generate different videos by sampling different layouts from the model. It also allows video manipulation through layout, where the user may remove an object, change the size and position of the objects, \etc.

\topic{User study}
We also conduct a user study to augment the quantitative evaluation. In the study, we present two videos generated by different methods together with the ground truth video and ask the raters 1) which video has the better visual quality, and 2) which video better reproduces the content of the ground truth video. We conduct the study with 40 videos and 11 participants. The results are in \cref{tab:study}, which is consistent with the quantitative results and further validates the superior performance of our method

\begin{table}[t]
    \small
    \centering
    \caption{\textbf{User study.} We report the percentage of raters that consider our method generates better video quality and better reproduces the ground truth videos respectively.
    }
    \vspace{-6pt}
    \label{tab:study}     
    \begin{tabular}{lcc}
        \toprule
        Methods & Quality & Reproduction\\
        \hline
        Ours \vs MAGVIT (frame pred.) &   76.3\% &  82.1\%  \\
        Ours \vs MAGVIT (class cond.) &  68.4\% &  66.7\% \\
        \bottomrule
    \end{tabular}
\end{table}

\begin{table}[t]
    \small
    \centering
    \caption{\textbf{Quantitative results for keyframe generation.} The metrics are averaged across keyframes. Please refer to the supplementary material for additional metrics. 
    }
    \vspace{-6pt}
    \label{tab:keyframe}     
    \begin{tabular}{lcccc}
        \toprule
        & \multicolumn{2}{c}{CoDraw} & \multicolumn{2}{c}{EPIC Kitchen}\\
        \cmidrule(lr){2-3} \cmidrule(lr){4-5}
        Methods & FID~$\downarrow$ & LPIPS~$\downarrow$ & FID~$\downarrow$ & LPIPS~$\downarrow$ \\
        \hline
        MaskGIT~\cite{chang2022maskgit} & 10.8 & 0.304 & 46.9 & 0.633 \\
        HCSS~\cite{jahn2021hcss} & 9.8  & 0.425 & 50.2 & 0.653 \\
        \hline
        Ours & 7.4  & 0.325 & 34.8 & 0.575 \\
        Ours-GT & 3.9  & 0.106 & 24.2 & 0.416 \\
        Ours-GT (Single) & 4.6 & 0.156 & 27.5 & 0.480 \\
        \bottomrule
    \end{tabular}
    \vspace{-12pt}
\end{table}

\begin{figure*}[t]
    \centering
    \includegraphics[width=1.0\textwidth]{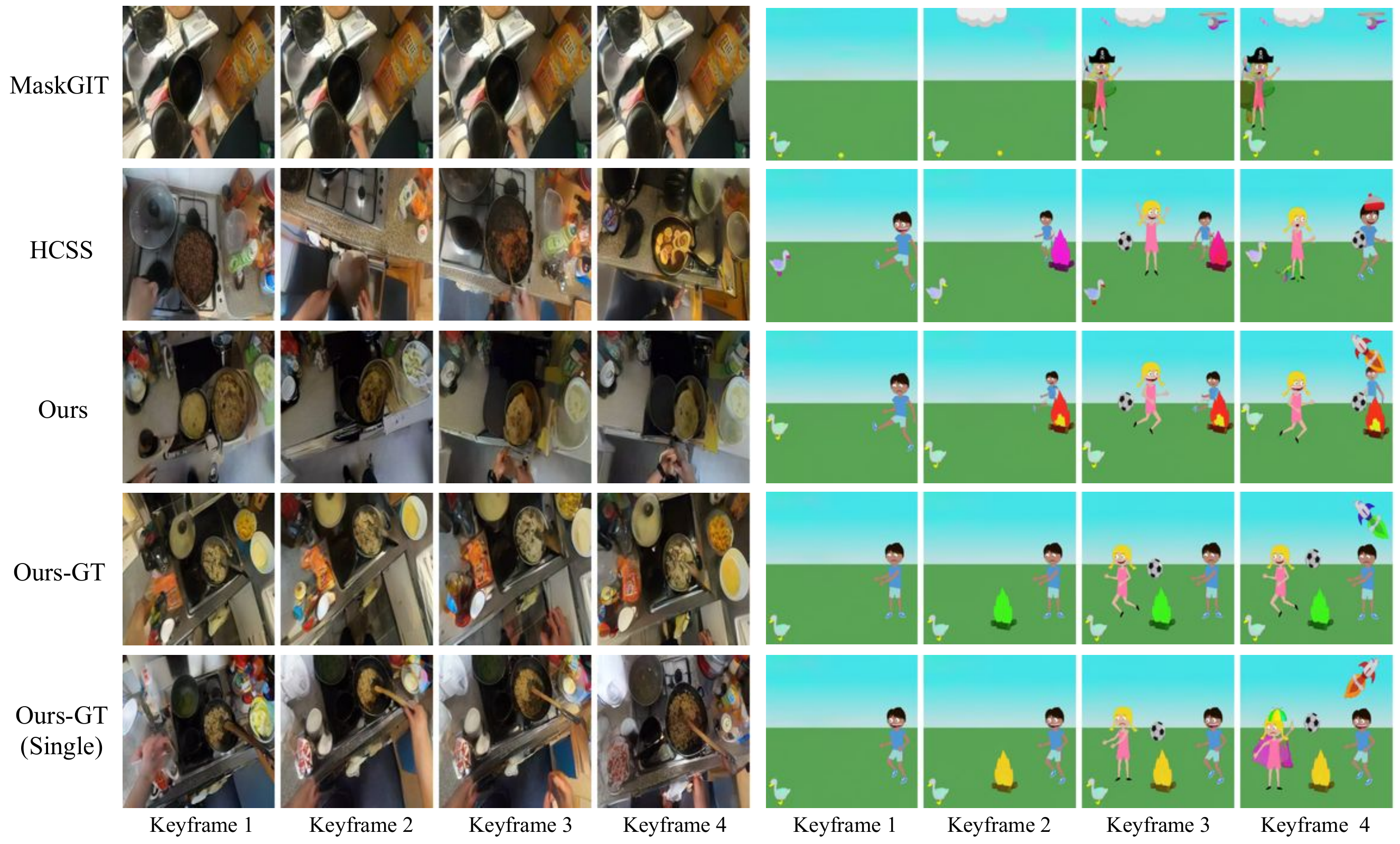}
    \vspace{-18pt}
    \caption{\textbf{Qualitative results for keyframe generation.} Compared with our method, MaskGIT tends to predict repetitive content which shows the importance of guidance. On the other hand, HCSS predicts inconsistent keyframes and fails to maintain temporal consistency. 
    When we compare Ours-GT with Ours-GT (Single), we can see the iterative approach fails to generate consistent keyframes. 
    The results show the importance of modeling the entire video jointly.
    }
    \label{fig:frame}
    \vspace{-9pt}
\end{figure*}

\subsection{Keyframes Generation}
\label{sub:keyframe}

Next, we evaluate the performance of our keyframe generation model. The goal is to verify 1) the importance of generating all the keyframes jointly, and 2) the importance of additional guidance for generating content across a large temporal window. 

\topic{Baselines}
We compare the following baselines and variants of our method:
\begin{itemize}[leftmargin=*,label=$\bullet$,topsep=2pt]
    \setlength{\itemsep}{2pt}
    \setlength{\parskip}{2pt}
    \item MaskGIT~\cite{chang2022maskgit}: the model takes the reference as input and iteratively predicts the next keyframe. This model represents keyframe generation without input guidance. 
    \item HCSS~\cite{jahn2021hcss}: the model takes a single layout as input and generates a single keyframe. We apply HCSS to generate each keyframe independently from the predicted layouts. This model represents keyframe generation without full video modeling.
    \item Ours: our keyframe generation given the predicted layouts as inputs.
    \item Ours-GT: our keyframe generation using the ground truth layouts as inputs (upper bound performance).
    \item Ours-GT (Single): our keyframe generator that predicts keyframes iteratively conditioning on the previous keyframe and ground truth layouts. This model represents keyframe generation without full video modeling.
\end{itemize}

\topic{Quantitative results} The results are in \cref{tab:keyframe}. Our method consistently outperforms HCSS on both real and synthetic data, which verifies the importance of joint prediction for all keyframes. Our method also performs better than MaskGIT except for the perceptual similarity with ground truth frames in the synthetic dataset. After taking a closer look at the generated frames, we observed that MaskGIT tends to predict repetitive keyframes with little changes across frames. This implies that the video will remain static, which is not suitable for video generation. These results show that our method can generate better keyframes for video generation than MaskGIT, which verifies the importance of guidance. We also compare different variants of our method. In particular, the superior performance of Ours-GT over Ours-GT (Single) further verifies that a model that considers the entire video jointly leads to better keyframe generation.

\topic{Qualitative results}
\cref{fig:frame} shows the qualitative results. As mentioned before, MaskGIT tends to generate keyframes with similar content, which shows the importance of providing input guidance at multiple timesteps. Comparing HCSS and our method, we can see that HCSS fails to generate consistent results across the keyframes, e.g.~the color of the fire changes. 
Similarly, when we compare Ours-GT with Ours-GT (Single), we can see the iterative approach fails to generate consistent keyframes. The examples clearly demonstrate the importance of joint modeling for the entire video.
Please refer to the supplementary material for additional evaluation, including the evaluation for layout generation.

\vspace*{-0.01in}
\section{Conclusions}
\vspace*{-0.01in}
\label{sec:conclusions}

We tackle the problem of long video generation which aims to generate videos beyond the output length of video generation models. We show that the existing sliding window approach is sub-optimal, and there is significant room for improvement using the same video generation model. To improve long video generation, we propose to use additional guidance to control the generation process. We further propose a two-stage approach which can utilize existing video generation models while capturing long-range dependency within the video. Empirical results validate our model design and show favorable results over state-of-the-art video generation methods.

{
\small
\bibliographystyle{ieee_fullname}
\bibliography{egbib}
}

\end{document}